\documentclass[10pt,twocolumn,letterpaper]{article}

\usepackage[pagenumbers]{cvpr} 

\definecolor{cvprblue}{rgb}{0.21,0.49,0.74}
\usepackage[pagebackref,breaklinks,colorlinks,allcolors=cvprblue]{hyperref}

\usepackage{listings}
\usepackage{epsfig}
\usepackage{amsmath}
\usepackage{amssymb}
\usepackage{verbatim}
\usepackage{booktabs}
\usepackage{tabularx}
\usepackage{amsfonts}
\usepackage{multirow}
\usepackage{bbding}
\usepackage{algorithm,algorithmicx,algpseudocode}
\usepackage{array}
\usepackage{colortbl}
\usepackage{natbib}
\usepackage{wasysym}
\usepackage{wrapfig}
\usepackage{pifont}
\usepackage{hhline}
\usepackage{float}
\usepackage{graphicx}

\definecolor{mygray}{gray}{.92}
\definecolor{lightgray}{gray}{.96}
\definecolor{myy}{RGB}{126,95,0}
\definecolor{ggray}{RGB}{127,127,127}
\definecolor{mygreen}{RGB}{0,0,0}
\definecolor{myred}{RGB}{240,16,89}
\definecolor{myblue}{RGB}{0,114,188}
\definecolor{darkgreen}{rgb}{0.0, 0.5, 0.0}
\definecolor{demphcolor}{RGB}{100,100,100}

\title{Mixed Degradation Image Restoration via Local Dynamic Optimization and Conditional Embedding}

\author{%
\bf \textbf{Yubin Gu}$^{1}$
\quad
\textbf{Yuan Meng}$^{1}$ \quad
\textbf{Xiaoshuai Sun}$^{1}$ \quad 
\textbf{Jiayi Ji}$^1$ \quad
\textbf{Weijian Ruan}$^2$ \\
\textbf{Rongrong Ji}$^1$ \\
\quad \vspace{1.em} \\
$^1$Key Laboratory of Multimedia Trusted Perception and Efficient Computing, Xiamen University \\
$^2$Smart City Research Institute, China Electronics Technology Group Corporation
}

\begin{document}
\twocolumn[{%
\renewcommand\twocolumn[1][]{#1}%
\maketitle

\begin{center}
    \captionsetup{type=figure}
    \includegraphics[width=1\textwidth]{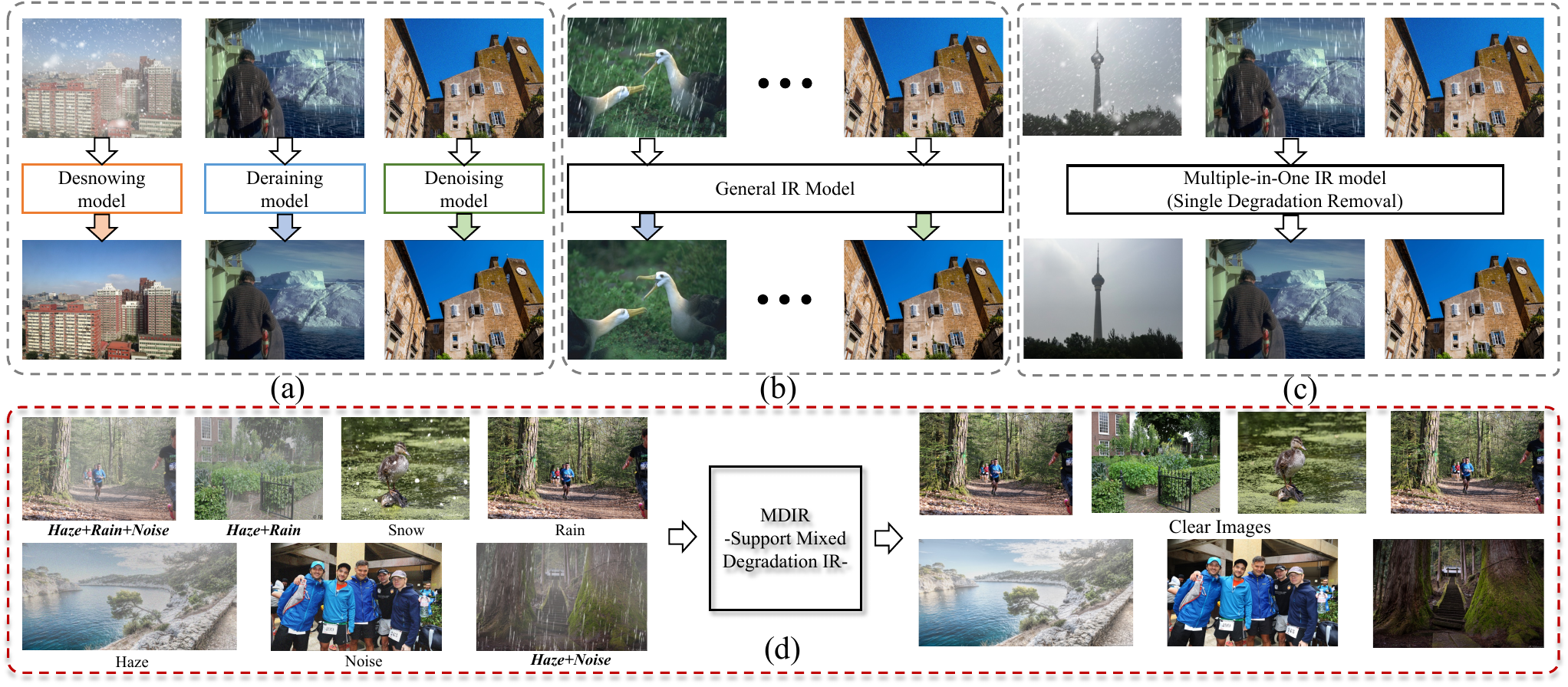}
    \captionof{figure}{(a) Models for specific single-degradation tasks require multiple models and training for different single degradations. (b) A general model for single-degradation restoration needs multiple training sessions for various single degradations. (c) A multi-task model for single-degradation restoration requires one model and one training session but lacks design for mixed-degradation scenarios. (d) Our proposed scenario is one in which a single model and training session can handle both single-degradation and mixed-degradation images.}
    \label{fig:inro-multipeDe}
\end{center}
}]

\begin{abstract}

Multiple-in-one image restoration (IR) has made significant progress, aiming to handle all types of single degraded image restoration with a single model. However, in real-world scenarios, images often suffer from combinations of multiple degradation factors. Existing multiple-in-one IR models encounter challenges related to degradation diversity and prompt singularity when addressing this issue. In this paper, we propose a novel multiple-in-one IR model that can effectively restore images with both single and mixed degradations. To address degradation diversity, we design a Local Dynamic Optimization (LDO) module which dynamically processes degraded areas of varying types and granularities. To tackle the prompt singularity issue, we develop an efficient Conditional Feature Embedding (CFE) module that guides the decoder in leveraging degradation-type-related features, significantly improving the model's performance in mixed degradation restoration scenarios. To validate the effectiveness of our model, we introduce a new dataset containing both single and mixed degradation elements. Experimental results demonstrate that our proposed model achieves state-of-the-art (SOTA) performance not only on mixed degradation tasks but also on classic single-task restoration benchmarks.
\end{abstract}    
\section{Introduction}
\label{sec:intro}
In real-world scenarios, various factors such as rain, snow, noise, and haze often cause significant degradation of captured images. Image restoration (IR), a crucial area of computer vision, aims to process these degraded images and restore high-quality outputs. Effective image restoration enhances the performance of downstream tasks, with applications spanning diverse fields such as remote sensing and autonomous driving.

IR is inherently an ill-posed problem, and traditional hand-crafted feature computation methods struggle to address the complexities of real-world environments. Recently, deep learning techniques have significantly advanced IR performance for specific degradation types—such as deraining~\cite{zhou2024uc,jiang2024fmrnet,chen2023hybrid,lin2024nightrain}, dehazing~\cite{chen2024dea,10005621}, denoising~\cite{herbreteau2024linear,flepp2024real}, and desnowing~\cite{chen2022snowformer}—by leveraging deep network models that are fine-tuned for each degradation type and learn effective priors from large datasets (Fig.~\ref{fig:inro-multipeDe}~(a)). However, most existing methods are designed to address only a single degradation type and lack robustness when confronted with unknown or combined degradations. Furthermore, some approaches~\cite{cui2024revitalizing,zamir2022restormer,wang2023promptrestorer} employ general models (Fig.~\ref{fig:inro-multipeDe}~(b)) and adjust model hyperparameters to target specific degradations, training the model individually for each case. While this reduces the need for manual design, it still requires selecting and fine-tuning models for unknown or mixed degradation, limiting its general applicability.

Recent efforts in multiple-in-one (MIO) IR models~\cite{potlapalli2024promptir,li2022all,cao2024hair} have aimed to address this gap by designing models capable of handling multiple single-degradation types (Fig.~\ref{fig:inro-multipeDe}~(c)). Early works, like AirNet~\cite{li2022all}, introduce contrastive learning based on image content in encoder features, improving the restoration of various degradations. However, this method struggles to disentangle different degradation types, particularly when multiple degradations coexist within the same scene. PromptIR~\cite{potlapalli2024promptir} further advances this concept by incorporating a degradation prompt module to guide feature expressions in the decoder, leveraging multi-head self-attention in a Transformer-based architecture. Despite its improvements, Transformer-based models face inefficiencies that hinder their performance. Challenges remain in addressing mixed degradation scenarios, such as those depicted in Fig.~\ref{fig:inro-multipeDe} (d), where the combination of multiple degradation types complicates restoration. Therefore, there is an urgent need for methods that can effectively restore images under such realistic and complex conditions.

Solving this task is inherently challenging due to two main obstacles. (1) Degradation diversity: Mixed degradation types introduce complex visual distortions, with varying degradation types and intensities across different spatial regions. This necessitates sophisticated handling of both high-frequency and low-frequency information at diverse locations, which existing methods are ill-equipped to address, leading to suboptimal restoration performance. (2) Prompt singularity: Although some multiple-in-one (MIO) methods incorporate specific prompts for individual degradations during the decoding stage, these prompts fail to effectively handle composite degradation types, failing when confronted with such scenarios.

To overcome these challenges, we propose a novel MIO image restoration model, MDIR, capable of effectively restoring both single and mixed degradation images, as shown in Fig.~\ref{fig:inro-multipeDe} (d). 
To address the degradation diversity issue, we introduce a Local Dynamic Optimization (LDO) module. This module dynamically generates a set of filters with varying categories and strengths based on input features, allowing the model to optimize high-frequency and low-frequency information across different image regions. 
To resolve the prompt singularity issue, we present a Conditional Feature Embedding (CFE) module with multi-label attributes. By pretraining on a mixed degradation dataset and fine-tuning through multi-label classification~\cite{yu2021multi} within a classification network, this module assigns multiple single-degradation labels to compound degraded images, enhancing the optimization process during decoding.

Furthermore, existing MIO methods often rely on datasets created by combining multiple single-task datasets. These datasets frequently suffer from imbalanced data distributions across degradation types and inconsistent scene conditions, which complicates the reliable evaluation of model performance. To address this, we introduce a new benchmark dataset, CIR, which includes seven mixed degradation types: rain, snow, haze, noise, rain+haze, haze+noise, and rain+haze+noise. The primary advantage of CIR is its balanced data distribution and the inclusion of images degraded under multiple conditions, enabling more accurate and consistent evaluation of MIO models. On the CIR dataset, our proposed MDIR outperforms the SOTA model PromptIR~\cite{potlapalli2024promptir} by over 1 dB in PSNR and demonstrates superior inference speed. Additionally, we achieve leading performance on three public single-degradation datasets: CSD~\cite{chen2021all_CSD} (desnowing) and Rain100L/H~\cite{yang2017deep_rain100LH} (deraining).

In summary, the primary contributions of this paper are as follows:
\begin{itemize}
\item We introduce a novel setting, Mixed Degradation Image Restoration, and propose an innovative IR model, MDIR, capable of handling both single and combined degradation types with a single model and checkpoint. 
\item In MDIR, the Local Dynamic Optimization Module (LDO) optimizes high- and low-frequency information across different regions in an input-adaptive manner, while the Conditional Feature Embedding (CFE) module effectively assists the decoder in identifying specific degradation types.
\item We construct a new multi-degradation dataset that provides a balanced distribution of degradation types within the same scene, establishing a novel benchmark for the IR community. Our method achieves state-of-the-art (SOTA) performance on both the proposed mixed degradation dataset and traditional IR datasets.

\end{itemize}


\section{Related Work}
\label{sec:related}
\begin{figure*}[t]
    \centering
    \includegraphics[width=1\linewidth]{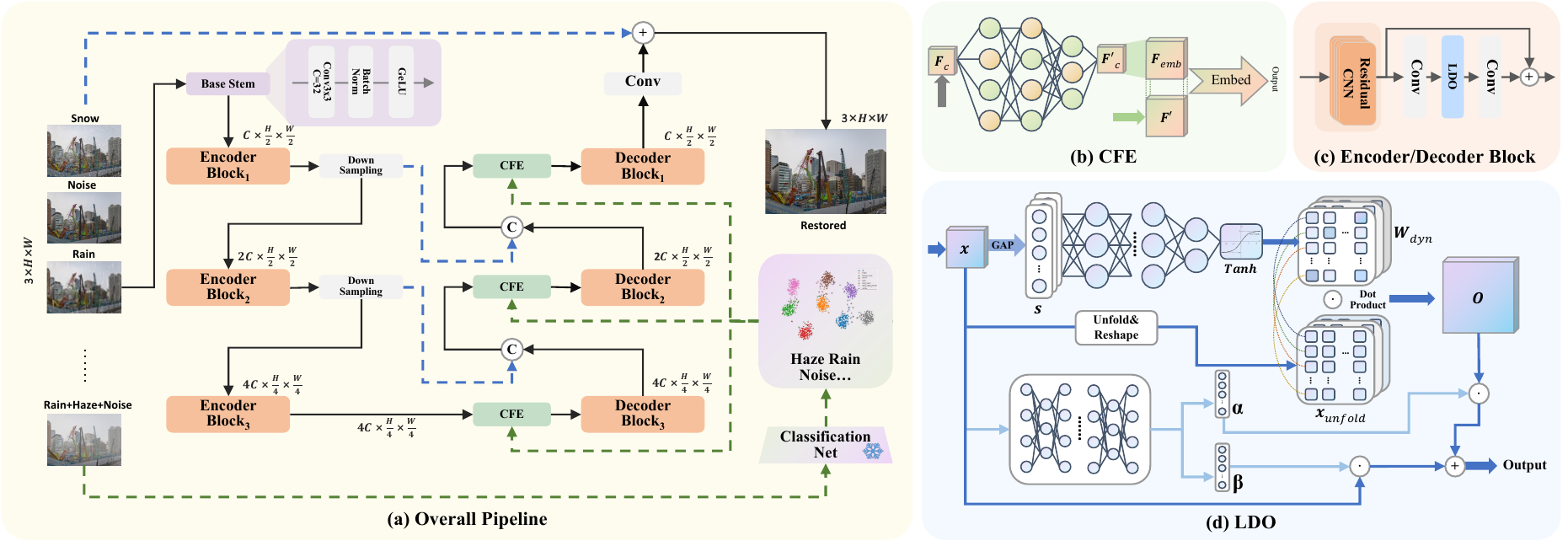}
    \caption{(a) illustrates the comprehensive architectural schematic of our proposed model, MDIR, adhering to the encoder-decoder paradigm. The encoding phase encompasses three sequential encoding blocks, while the decoding phase comprises three decoding blocks along with three instances of the Conditional Feature Embedding Module (CFE). The configuration of each encoding and decoding block is delineated in (c), showcasing an assembly of multiple residual blocks succeeded by the LDO-cored structure (d). Sub-figure (b) denotes the CFE, which facilitates the integration between the classifier and the primary restoration network.
}
    \label{fig:framework}
\end{figure*}
Image restoration (IR) constitutes a pivotal research area within the field of computer vision. Vanilla approaches based on hand-crafted feature design exhibit limited efficacy when confronted with complex and variable degradation scenarios. With the rapid advancement of deep learning, the domain of IR has experienced significant progress. Data-driven methodologies are capable of more comprehensively elucidating degradation mechanisms and delivering more effective restoration outcomes. In recent years, numerous IR techniques targeting specific types of degradation—such as dehazing~\cite{chen2024dea,10005621}, deraining~\cite{zhou2024uc,jiang2024fmrnet}, denoising~\cite{herbreteau2024linear}, and desnowing~\cite{chen2022snowformer}—have emerged. These methods typically leverage CNNs or Transformer architectures, each presenting distinct advantages and limitations. For instance, CNN-based approaches~\cite{cui2024revitalizing,zamir2021multi_mpr} are highly efficient in processing image data, offer rapid inference speeds, and facilitate quicker convergence during training. Conversely, Transformer-based methods~\cite{chen2022snowformer,zamir2022restormer,zhou2024uc_ucf} excel at capturing long-range feature dependencies and demonstrate superior performance on large-scale datasets; however, their quadratic computational complexity in attention mechanisms poses significant constraints on their applicability. While these degradation-specific designs achieve remarkable results within their respective application domains, their specialized nature leads to increased costs in model design, training, and deployment, particularly in environments where multiple degradation types coexist, such as "Haze+Rain", "Haze+Noise" and "Haze+Noise+Rain" as shown in the left side on Fig.~\ref{fig:inro-multipeDe}~(d).

To mitigate design costs, researchers have developed general-purpose IR models~\cite{cui2023image,zamir2022restormer,lin2024improving_zuo}, such as IRNeXt~\cite{cui2024revitalizing}, Restormer~\cite{zamir2022restormer}, and PromptRestorer~\cite{wang2023promptrestorer}. These models are not tailored to specific tasks, thereby substantially reducing development expenses while still delivering satisfactory performance across various degradation types. Nevertheless, these universal models require hyperparameter adjustments and individual training sessions for different tasks. When dealing with images afflicted by multiple degradation types, the necessity to prepare multiple sets of training parameters further escalates deployment costs. Recently, multi-in-one IR approaches~\cite{li2022all,potlapalli2024promptir,lin2024improving_zuo} have been recognized as effective solutions to address the aforementioned challenges. For example, Li et al.~\cite{li2022all} employed content-based contrastive learning to differentiate encoded features, thereby guiding the decoder to autonomously learn restoration processes for different degradations. Potlapalli et al.~\cite{potlapalli2024promptir} introduced PromptIR, which incorporates a series of learnable parameters to supply degradation-related information, subsequently injecting this information into the decoding process. While these methods have demonstrated excellent performance in specific scenarios, they have yet to address situations where multiple degradations coexist within a single image. Additionally, the complexity of their discrimination mechanisms contributes to increased model complexity. In this study, we not only consider scenarios where multiple single degradations coexist but also address cases where a single image is subjected to multiple simultaneous degradations. We have meticulously designed both the primary restoration model and the prompt-guided mechanism, thereby providing a novel benchmark model for this challenging problem.

\section{Methods}
\label{sec:methods}

In this section, we provide a comprehensive exposition of our proposed model, MDIR, a novel multiple-in-one image restoration approach. 

\subsection{Overall Pipeline}
We introduce MDIR, a model that restores a clear image \( \mathbf{I}_{\text{clear}} \) from a degraded input image \( \mathbf{I}_{\text{degra}} \) without prior knowledge of the degradation type. The MDIR comprises two main components: the primary restoration network, responsible for image restoration, and the degradation classification network, which extracts semantic features with label identities and feeds them into the conditional feature embedding module (CFE) to guide decoding.

The model accepts an RGB degraded image $\mathbf{I}_{\text{degra}} \in \mathbb{R}^{3 \times H \times W}$ and adopts BaseStem block (Fig.~\ref{fig:framework}~(a), pink block) to extract shallow features $\mathbf{\bar{I}} \in \mathbb{R}^{C \times H \times W}$ using channel expansion convolution layers, where $C = 32$. The degraded image is also processed by a pre-trained multi-label classification network to generate semantic features with label identities for the CFE (Fig.~\ref{fig:framework}~(b)). During encoding, the input features pass through three encoding blocks, producing features at multiple scales: $\mathbf{\bar{I}}_e = \{ \mathbb{R}^{2^{i-1} C \times \tfrac{H}{2^{i-1}} \times \tfrac{W}{2^{i-1}}} \mid i = 1, 2, 3 \}$. These features proceed to the decoding stage. Before each decoding layer, the CFE enriches the features with label identities and semantic information. Except for the final decoding block, each decoding block receives input formed by concatenating the output from the previous decoding block with the corresponding encoding features. Each encoding block processes features through seven convolutional residual structures and refines them using a local dynamic optimization module (LDO). The output of each decoding block is passed to the next and also used to generate a predicted clear image at that layer, implementing deep supervision to accelerate learning. We will detail the design and function of the LDO and the CFE in the next two sub-sections.
\vspace{0.5em}
\subsection{Local Dynamic Optimization Module}
In a degraded image, different regions often suffer varying degrees of degradation, particularly in scenarios where multiple degradation types coexist. Vanilla convolutional neural networks (CNNs) with static filters struggle to effectively handle spatially varying degradations due to their fixed weights and translational invariance. Although Transformer models with self-attention mechanisms excel at capturing long-range dependencies, their quadratic computational complexity renders them inefficient. To tackle this dilemma, we propose a Local Dynamic Optimization module (LDO) (see Figure~\ref{fig:framework}~(d)), which combines the advantages of convolutional operations and attention mechanisms while mitigating their respective drawbacks, making it particularly suitable for complex IR tasks.

Given an input feature map $\mathbf{x} \in \mathbb{R}^{C \times H \times W}$, where $C$ denotes the number of channels and $H$ and $W$ represent the height and width respectively, we first apply global average pooling (\(GAP(\cdot)\)) to preliminarily extract the global context of the input features:
\begin{equation}
\mathbf{s} = GAP(\mathbf{x}).    
\end{equation}

Next, global abstract feature \( \mathbf{s} \) is fed into a weight generation network to produce dynamic convolutional kernels. In other words, a set of filters is generated based on the feature space dimension. This weight generation network comprises two \(1 \times 1\) convolutional layers followed by ReLU functions, formulated as:

\begin{equation}
\mathbf{s'} = ReLU\left( BN({Conv}_{1 \times 1}(\mathbf{s}) )\right), \quad \mathbf{s'} \in \mathbb{R}^{\frac{C}{r} \times 1 \times 1},
\end{equation}

\begin{equation}
\mathbf{w} =  {Conv}_{1 \times 1}(\mathbf{s'}),
\end{equation}
\begin{equation}
    \mathbf{W_{\text{dyn}}} =  \frac{e^\mathbf{w} - e^{-\mathbf{w}}}{e^\mathbf{w} + e^{-\mathbf{w}}}, \quad \mathbf{W_{\text{dyn}}}\in \mathbb{R}^{(C \times k^2) \times 1 \times 1},
\end{equation}
where \( r \) is the channel reduction factor and \( k \) is the kernel size. We employ the hyperbolic tangent function to normalize the value ranging within \((-1, 1)\), to avoid the potential single low-pass effect caused by softmax or sigmoid functions. This strategy allows for greater flexibility in addressing the optimization needs for high and low frequency information in different local regions.
We then reshape the weight parameters to form the dynamic convolutional kernels:

\begin{equation}
\mathbf{W_{\text{dyn}}} = reshape(\mathbf{{W}_{\text{dyn}}}), \quad  \mathbf{W_{\text{dyn}}} \in \mathbb{R}^{(C \times k^2)}.
\end{equation}

After obtaining the dynamic convolutional kernels \(\mathbf{W}_{\text{dyn}}\), the input feature \( \mathbf{x} \) is unfolded to facilitate element-wise multiplication with the corresponding convolutional kernels:

\begin{equation}
\mathbf{X}_{\text{unfold}} = {unfold}(\mathbf{x}, k), \quad \mathbf{X}_{\text{unfold}} \in \mathbb{R}^{C \times k^2 \times (H \times W)}.
\end{equation}

We then perform element-wise multiplication and summation to obtain the feature map \( \mathbf{O} \), enriched with high-frequency information:

\begin{equation}
\mathbf{O} =  \sum_{i=1}^{k^2} W_{\text{dyn}} \cdot X_{\text{unfold}}, \quad \mathbf{O} \in \mathbb{R}^{C \times H \times W},
\end{equation}

Since the original input feature \( \mathbf{x} \) contains more low-frequency information—which is crucial for restoring smooth regions in degraded images—we perform a weighted fusion of \( \mathbf{x} \) and \( \mathbf{O} \) to obtain the final output feature map:

\begin{equation}
\mathbf{F}_{\text{out}} = \alpha \cdot \mathbf{O} + \beta \cdot \mathbf{x}, \quad \mathbf{F}_{\text{out}} \in \mathbb{R}^{C \times H \times W},
\end{equation} where \( \alpha \) and \( \beta \) are learnable parameters dependent on the input \( x \), expressed as:

\begin{equation}
[\alpha, \beta] = {sigmoid}\left( MLP(\mathbf{x}) \right).
\end{equation}Here, the use of the sigmoid function ensures the weight values in \([0,1]\).

Through this design, LDO adaptively balances high- and low-frequency information, enhancing the model's capability to handle spatially varying degradations effectively.

\subsection{Conditional Feature Embedding Module}
In multiple-in-one image restoration models, distinguishing features associated with different degradation types is an effective means to enhance performance. That is to say, an effective signal can guide the decoder to optimize for different restoration tasks. Unlike previous approaches that rely on contrastive learning or classification networks trained on encoded features, this study introduces a lightweight classification network to guide the decoding process. Inspired by the positional encoding in Transformers, we designed a simple yet efficient conditional feature embedding module (CFE), as illustrated in Fig.~\ref{fig:framework}. The CFE provides semantic identifiers for the decoded features, enabling the decoder to effectively differentiate among various degradation types and thus enhance decoding optimization.

Specifically, the degraded image \( \mathbf{I}_{\text{degra}} \in \mathbb{R}^{3 \times H \times W} \) is input into a classification network, in this case a fine-tuned MobileNetv2~\cite{sandler2018mobilenetv2} model pre-trained for multi-label classification, to efficiently extract semantic features. The extracted features are then adjusted to match the channel dimensions via a \(1 \times 1\) convolution layer, followed by the bilinear interpolation to achieve spatial alignment. This process yields the conditional embedding features:

\begin{equation}
    \mathbf{F}_{c} = {Conv}_{1 \times 1}({MobileNet}(\mathbf{I}_{\text{degra}}))    
\end{equation}

\begin{equation}
    \mathbf{F}_{\text{embedding}} = {Upsampling}(\mathbf{F}_c)    
\end{equation}

Next, the output features from the previous decoding layer and the corresponding encoding layer features are concatenated along the channel dimension, forming the initial input:

\begin{equation}
    \mathbf{F}' = {Conv}_{1 \times 1}({Concat}(\mathbf{\bar{I}_e^{i}}, \mathbf{I}_d^{i-1}))    
\end{equation}

Then the conditional embedding features are incorporated into \( \mathbf{F'} \), producing the input for the current decoding block:
\begin{equation}
\mathbf{F}_d^{\text{in}} = \mathbf{F}' + \mathbf{F}_{\text{embedding}}    
\end{equation}

Through the processing of the CFE, the input features are enriched with class-specific and semantic identifiers, enabling the model to distinguish and process features according to the degradation type.

\subsection{Loss Function}
Our model comprises two main components: a primary restoration network and a multi-label classification network. Each component undergoes independent training to optimize its respective function. We employ a fine-tuned, pre-trained MobileNet model for the multi-label classification network, utilizing binary cross-entropy as the loss function. During fine-tuning, we represent hybrid degradation classes as combining multiple independent classes rather than a single unified category (\emph{i.e.}, non-one-hot encoding). This approach enables the hybrid classes to retain individual degradation characteristics. The primary restoration network's decoding stage outputs predictions at three hierarchical levels, with the total loss function defined as the cumulative loss across these three stages. Drawing from previous image restoration methods, we calculate the L1 loss between the predicted and ground truth images in both spatial and Fourier domains, using a weighted sum of these values as the final loss. This dual-domain approach captures both spatial and frequency-domain features of the image, enhancing restoration fidelity.

\begin{table*}[!ht]
\caption{Comparison results of multiple types of degradation (including mixed degradation). "h+n" denotes the composite degradation of haze and noise, "r+h" indicates the composite of rain and haze, and "r+h+n" signifies the composite of rain, haze, and noise.}
\centering
\arrayrulecolor{black}
\setlength{\tabcolsep}{1.3pt}
\renewcommand{\arraystretch}{1.4}
\begin{tabular}{c|cc|cc|cc|cc|cc|cc|cc|cc} 
\specialrule{1.2pt}{0pt}{0pt}
\rowcolor[rgb]{1,1,1} {\cellcolor[rgb]{1,1,1}}                          & \multicolumn{2}{c|}{haze} & \multicolumn{2}{c|}{h+n} & \multicolumn{2}{c|}{noise} & \multicolumn{2}{c|}{rain} & \multicolumn{2}{c|}{r+h} & \multicolumn{2}{c|}{r+h+n} & \multicolumn{2}{c|}{snow} & \multicolumn{2}{c}{Averages}  \\ 
\hhline{>{\arrayrulecolor[rgb]{1,1,1}}->{\arrayrulecolor{black}}----------------}
\rowcolor[rgb]{1,1,1} \multirow{-2}{*}{{\cellcolor[rgb]{1,1,1}}Methods} & PSNR   & SSIM             & PSNR   & SSIM                    & PSNR   & SSIM              & PSNR   & SSIM             & PSNR   & SSIM                  & PSNR   & SSIM                        & PSNR   & SSIM             & PSNR   & SSIM                \\ 
\hline
AirNet                                                                                          & 15.295 & 0.446            & 15.770 & 0.452                   & 17.624 & 0.475             & 16.747 & 0.461            & 15.299 & 0.435                 & 15.827 & 0.446                       & 16.569 & 0.439            & 16.162 & 0.451               \\
PromptIR                                                                                        & 30.999 & 0.982            & 27.595 & 0.918                   & 35.224 & 0.952             & 39.573 & 0.990            & 30.794 & 0.973                 & 27.829 & 0.905                       & 32.914 & 0.953            & 32.133 & 0.953               \\
Restormer                                                                                       & 30.482 & 0.982            & 27.513 & 0.914                   & 35.100 & 0.950             & 38.561 & 0.988            & 30.055 & 0.969                 & 27.641 & 0.901                       & 31.995 & 0.946            & 31.621 & 0.950               \\
IRNext                                                                                          & 31.355 & \textbf{0.984}            & 27.728 & 0.924                   & 35.282 & 0.954             & 40.422 & 0.991            & 31.453 & 0.976                 & 28.070 & 0.914                       & 33.665 & 0.957            & 32.568 & 0.957               \\
AST                                                                                             & 28.251 & 0.979            & 26.025 & 0.912                   & 34.945 & 0.950             & 37.152 & 0.986            & 28.341 & 0.963                 & 26.233 & 0.897                       & 31.327 & 0.939            & 30.325 & 0.946               \\ 
\hline
\rowcolor[rgb]{0.851,0.851,0.851} Ours                                                          & \textbf{31.687} & \textbf{0.984}            & \textbf{28.571} & \textbf{0.928}                   & \textbf{35.499} & \textbf{0.956}             & \textbf{41.620} & \textbf{0.993}            & \textbf{31.847} & \textbf{0.978}                 & \textbf{28.458} & \textbf{0.917}                       & \textbf{34.811} & \textbf{0.964}            & \textbf{33.213} & \textbf{0.960}               \\
\specialrule{1.2pt}{0pt}{0pt}
\end{tabular}
\label{tab:mixedtype}
\arrayrulecolor{black}
\end{table*}
\begin{figure*}
    \centering
    \includegraphics[width=1\linewidth]{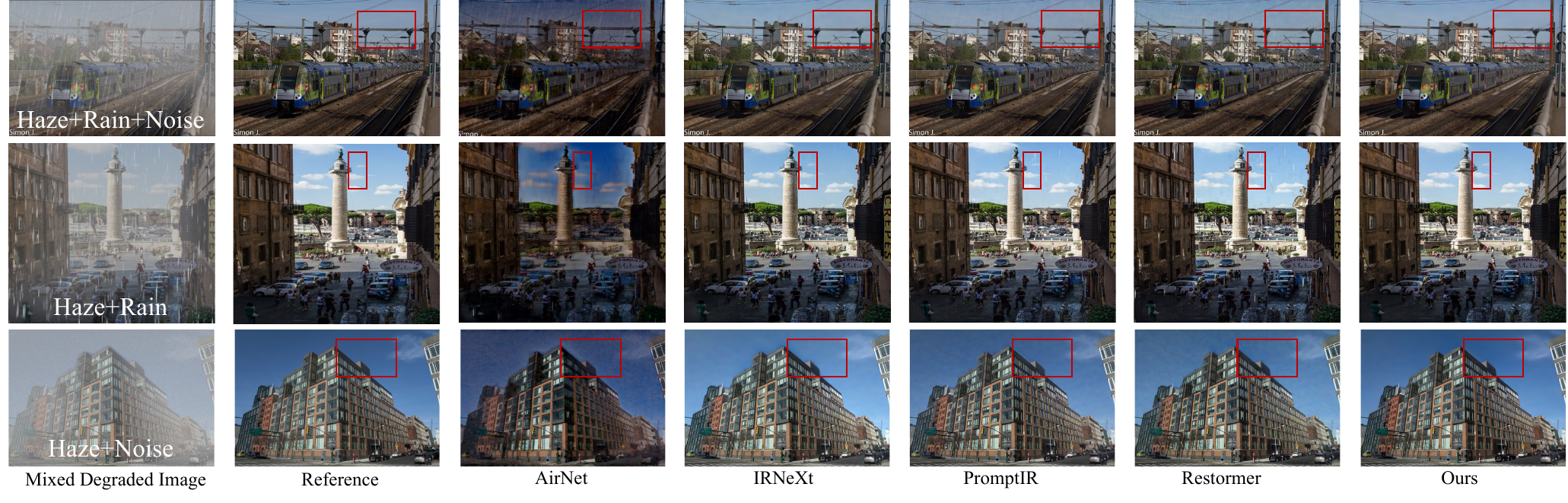}
    \caption{The results of various methods for mixed degradation image restoration.}
    \label{fig:mixeddeg}
\end{figure*}

\section{Experiments}
\label{sec:experiments}
This section details our experiments, including implementation specifics, datasets, performance on mixed degradation and single-task datasets, and ablation studies.
\subsection{Implementation Details}
Our proposed model comprises two main components: a restoration network and a multi-label classification network, which is based on the efficient MobileNetv2 model. Accordingly, we employ a two-stage training strategy. First, the multi-label classification network is fine-tuned; initialized with pre-trained parameters, it requires only 5 epochs to reach convergence, incurring minimal training overhead. To train the restoration network, we freeze the parameters of the classification network. The restoration network follows an end-to-end training approach. Both the encoder and decoder contain three stages. Each encoding block includes seven consecutive residual structures, followed by a multi-scale residual structure centered on LDO. The configuration of each decoding block mirrors that of its corresponding encoding block.

\begin{table}[!t]
\centering
\setlength{\tabcolsep}{1.2pt}
\caption{Comparison of the number of parameters and FLOPs.}
\begin{tabular}{ccccccc} 
\hline
Methods  & AirNet & PromptIR & Restormer & IRNext & AST   & Ours  \\ 
\hline
Para.(M) & 9.0    & 33.0     & 24.5      & 13.2   & 110.3 & 8.9   \\
FLOPs(G) & 77.9   & 158.1    & 174.7     & 114.0  & 65.3  & 71.5  \\
Speed(s) & 0.039   & 0.088    & 0.083     & 0.025  & 0.065  & 0.026  \\
\hline
\end{tabular}
\label{tab:para}
\end{table}
\begin{table*}[h]
\caption{
The results were obtained from unified training of multiple degradations and one-to-one training of multiple degradations on CIR. "$\text{Ours}_{all}$" represents the unified training strategy, and "$\text{Ours}_{one}$" represents the one-to-one training strategy.}
\setlength{\tabcolsep}{3pt}
\renewcommand{\arraystretch}{1.4}
\centering
\begin{tabular}{c|cc|cc|cc|cc|cc|cc|cc} 
\specialrule{1.2pt}{0pt}{0pt}
\multirow{2}{*}{Methods} & \multicolumn{2}{c|}{haze} & \multicolumn{2}{c|}{haze+noise} & \multicolumn{2}{c|}{noise} & \multicolumn{2}{c|}{rain} & \multicolumn{2}{c|}{rain+haze} & \multicolumn{2}{c|}{rain+haze+noise} & \multicolumn{2}{c}{snow}  \\ 
\cline{2-15}
                         & PSNR   & SSIM             & PSNR   & SSIM                    & PSNR   & SSIM              & PSNR   & SSIM             & PSNR   & SSIM                   & PSNR   & SSIM                          & PSNR   & SSIM             \\ 
\hline
$\text{Ours}_{all}$                 & 31.687 & 0.984            & 28.571 & 0.928                   & 35.499 & 0.956             & 41.620 & 0.993            & 31.847 & 0.978                  & 28.458 & 0.917                         & 34.811 & 0.964            \\
$\text{Ours}_{one}$                 & 30.056 & 0.983            & 27.679 & 0.925                   & 35.736 & 0.958             & 42.276 & 0.994            & 31.283 & 0.978                  & 27.619 & 0.914                         & 35.583 & 0.969            \\
\specialrule{1.2pt}{0pt}{0pt}
\end{tabular}
\label{tab:unifiy_specific}
\end{table*}
\vspace{0.5em}
During training, the batch size is set to 16 (covering both mixed and single tasks), with a total of 800 epochs, using the Adam optimizer. The initial learning rate is $3 \times 10^{-4}$ and is gradually reduced to $1 \times 10^{-6}$ adopting the cosine annealing schedule. Training images are randomly cropped to $256 \times 256$ patches, while in testing, original full-size images are used without any cropping or patching strategy. The model is implemented in PyTorch and trained on eight NVIDIA 3090 GPUs (each with about 24 GB of memory).
\subsection{Datasets}
We evaluate our model on both a mixed degradation dataset and datasets tailored for specific restoration tasks.

\textbf{Mixed Degradation Dataset:} To validate our method’s capability in handling various single and hybrid degradation images within a single model, we constructed a comprehensive dataset, CIR, encompassing seven degradation types: rain, snow, haze, noise, rain + haze, haze + noise, and rain + haze + noise. CIR provides a variety of scenes and diverse image content. Additionally, the dataset includes images with identical content affected by different degradations, posing a more significant challenge for degradation identification and restoration. CIR contains 7,952 training images and 1,050 test images, with each degradation type represented by 1,136 training images and 150 test images. Competing methods were retrained based on the descriptions in their original papers until they reached convergence.

\textbf{Single Degradation IR Datasets:} We evaluated the generalization capability of our model on single degradation tasks, specifically deraining and desnowing. We utilized the widely adopted Rain100L~\cite{yang2017deep_rain100LH} and Rain100H datasets for deraining and the CSD~\cite{chen2021all_CSD} dataset for desnowing. Evaluation metrics for comparison methods are obtained from their original papers.

Following the evaluation metrics used by other IR methods, We also use Peak Signal-to-Noise Ratio (PSNR) and Structural Similarity Index Measure (SSIM) to quantitatively assess the model’s performance.

\subsection{Mixed Degradation IR Results}
Testing on mixed degradation datasets allows us to assess the performance of a multiple-in-one IR model across multiple degraded images. We evaluated our model on the CIR dataset, which includes seven categories: four single degradation types and three composite degradation types. Table~\ref{tab:mixedtype} presents a quantitative comparison of our method with five recent approaches, including two multiple-in-one restoration models, AirNet~\cite{li2022all} and PromptIR~\cite{potlapalli2024promptir}, and three general-purpose multi-task models, Restormer~\cite{zamir2022restormer}, IRNeXt~\cite{cui2024revitalizing}, and AST~\cite{zhou2024adapt_ast}. Results show that although AirNet performs well on datasets used in its original paper, its content-based contrastive learning strategy struggles to handle the CIR dataset’s challenges, where scenes with identical content are affected by different degradations, leading to performance degradation. Compared to the multiple-in-one method PromptIR, our model consistently outperforms across all tasks, achieving an average PSNR gain exceeding 1 dB. Additionally, our model exhibits clearer visual outputs, reflecting superior restoration quality. This advantage stems from MDIR’s local dynamic optimization strategy, as well as its conditional feature embedding mechanism. Specifically, LDO ensures high restoration performance, while CFE assists in distinguishing various degradation types, facilitating synergistic improvement across tasks. Table~\ref{tab:para} compares the number of parameters and FLOPs of each method, and the results show that MDIR still has an advantage.
\subsection{Single Degradation IR Results}
\subsubsection{Results of Single-task on CIR}
In addition to mixed training on the CIR dataset, we further investigated the performance of MDIR when individually trained on different degradation types within CIR, with results shown in Table ~\ref{tab:unifiy_specific}. It can be observed that in noise, rain, and snow scenarios, individually trained models yield better results; however, for haze (including mixed degradations), unified training enhances the model’s restoration performance. This suggests that mixed training does not necessarily result in performance loss across all scenarios but can have a beneficial effect on specific degradation types

\begin{table}[t]
\centering
\caption{Results of various methods on the single image deraining. The datasets are Rain100L and Rain100H, which contain light rain streaks and heavy rain streak degradation, respectively.}
\setlength{\tabcolsep}{6.6pt}
\renewcommand{\arraystretch}{1}
\arrayrulecolor{black}
\begin{tabular}{c|cc|cc} 
\specialrule{1.2pt}{0pt}{0pt}
\rowcolor[rgb]{1,1,1} {\cellcolor[rgb]{1,1,1}}                          & \multicolumn{2}{c|}{Rain100L} & \multicolumn{2}{c}{Rain100H}  \\ 
\hhline{>{\arrayrulecolor[rgb]{1,1,1}}->{\arrayrulecolor{black}}----}
\rowcolor[rgb]{1,1,1} \multirow{-2}{*}{{\cellcolor[rgb]{1,1,1}}Methods} & PSNR   & SSIM                 & PSNR   & SSIM                 \\ 
\hline
FSNet                                                                                           & 38.000 & 0.972                & 31.770 & 0.906                \\
HINet                                                                                           & 37.280 & 0.970                & 30.650 & 0.894                \\
MPRNet                                                                                          & 36.400 & 0.965                & 30.410 & 0.890                \\
Restormer                                                                                       & 38.990 & 0.983                & 30.820 & 0.896                \\
MFDNet                                                                                          & 37.610 & 0.973                & 30.480 & 0.899                \\
MambaIR                                                                                         & 38.780 & 0.977                & 30.620 & 0.893                \\
UC-former                                                                                       & 38.890 & 0.978                & 30.780 & 0.904                \\
PromptRestorer                                                                                  & 39.040 & 0.977                & 31.720 & 0.908                \\
\hline
\rowcolor[rgb]{0.851,0.851,0.851} Ours                                                          & \textbf{39.090} &\textbf{0.983}                & \textbf{33.086} & \textbf{0.925}                \\
\specialrule{1.2pt}{0pt}{0pt}
\end{tabular}
\label{tab:rain100LH}
\end{table}

\subsubsection{Deraining}
For the deraining task, we compared our method with eight recent advanced approaches, spanning CNN, Transformer, and Mamba architectures, including FSNet~\cite{cui2023image}, HINet~\cite{chen2021hinet_hin}, MPRNet~\cite{zamir2021multi_mpr}, Restormer~\cite{zamir2022restormer}, MFDNet~\cite{wang2023multi_mfd}, MambaIR~\cite{guo2025mambair_mamir}, UCFormer~\cite{zhou2024uc_ucf}, and PromptRestorer~\cite{wang2023promptrestorer}. Table ~\ref{tab:rain100LH} presents their performance on Rain100L~\cite{yang2017deep_rain100LH} and Rain100H~\cite{yang2017deep_rain100LH}. Our method achieves superior performance, particularly on Rain100H, where the PSNR surpasses the second-best method by over 1 dB, and SSIM is also higher. In scenarios with severe rain streaks, our model produces significantly clearer restored images.

\subsubsection{Desnowing}
In the single-image desnowing task, we compared our method with 9 advanced approaches, encompassing methods specifically designed for desnowing and general-purpose models, including IRNeXt~\cite{cui2024revitalizing}, Restormer~\cite{zamir2022restormer}, SnowFormer~\cite{chen2022snowformer}, FSNet~\cite{cui2023image}, TransWeather~\cite{valanarasu2022transweather_transw}, HCSD-Net~\cite{zhang2023hcsd_hcsd}, BMNet~\cite{chen2023uncertainty_bmnet}, OKNet~\cite{cui2024omni_oknet}, and PromptRestorer~\cite{wang2023promptrestorer}. Table~\ref{tab:snowcsd} presents the results of these methods on the CSD dataset~\cite{chen2021all_CSD}. Our model outperforms all others in both PSNR and SSIM metrics, notably surpassing the dedicated desnowing method SnowFormer, demonstrating the strong generalization capability of MDIR.

\begin{table}
\centering
\caption{Comparison results of various methods on the single image snow removal task. The dataset is CSD.}
\arrayrulecolor{black}
\setlength{\tabcolsep}{20pt}
\renewcommand{\arraystretch}{1.3}
\begin{tabular}{c|cc} 
\specialrule{1.2pt}{0pt}{0pt}
\rowcolor[rgb]{1,1,1} {\cellcolor[rgb]{1,1,1}}                          & \multicolumn{2}{c}{CSD}  \\ 
\hhline{>{\arrayrulecolor[rgb]{1,1,1}}->{\arrayrulecolor{black}}--}
\rowcolor[rgb]{1,1,1} \multirow{-2}{*}{{\cellcolor[rgb]{1,1,1}}Methods} & PSNR   & SSIM            \\ 
\hline
IRNext                                                                                          & 37.653 & 0.980           \\
Restormer                                                                                       & 35.430 & 0.973           \\
SnowFormer                                                                                      & 39.450 & 0.982           \\
FSNet                                                                                           & 38.370 & 0.986           \\
TransWeather                                                                                    & 31.760 & 0.932           \\
HCSD-Net                                                                                        & 33.529 & 0.963           \\
BMNet                                                                                           & 38.090 & 0.979           \\
OKNet                                                                                           & 37.990 & 0.985           \\
PromptRestorer                                                                                  & 37.480 & 0.985           \\
\hline
\rowcolor[rgb]{0.851,0.851,0.851} Ours                                                          & \textbf{39.772} & \textbf{0.986}           \\
\specialrule{1.2pt}{0pt}{0pt}
\end{tabular}
\label{tab:snowcsd}
\end{table}

\subsection{Ablation Study}
In this section, a more detailed ablation study is conducted on the proposed model MDIR, primarily to verify the impact of the LDO module and the CFE module on the model's performance.

\subsubsection{The Effectiveness of LDO and CFE}
To verify the gains brought by two key modules in MDIR, namely 
Local Dynamic Optimization Module (LDO) and Conditional Feature Embedding Module (CFE), we used a model without these two modules as the baseline. We then incrementally added LDO and CFE to the baseline model, and the results are shown in Table~\ref{tab:abla_module}. Specifically, compared to the baseline model, both LDO and CFE improved performance, with LDO showing a more significant gain.

Furthermore, we embedded CFE into two models designed for general restoration tasks, IRNeXt and Restormer. The specific results are shown in Table~\ref{tab:cfeabla}, where we can see that after being equipped with CFE, both models showed further improvement on average, further demonstrating the effectiveness of CFE.
\begin{table}
\centering
\caption{The ablation study results for the two key modules.}
\setlength{\tabcolsep}{11.8pt}
\begin{tabular}{c|cc} 
\specialrule{1.2pt}{0pt}{0pt}
\multirow{2}{*}{}     & \multicolumn{2}{c}{Average (CIR)}  \\ 
\cline{2-3}
                      & PSNR   & SSIM                      \\ 
\hline
Baseline              & 31.030 & 0.948                     \\
Baseline + LDO       & 32.937 & 0.958                     \\
Baseline + LDO + CFE & 33.213 & 0.960                     \\
\specialrule{1.2pt}{0pt}{0pt}
\end{tabular}
\label{tab:abla_module}
\end{table}

\begin{table}
\centering
\caption{The impact results of CFE on other two models.}
\setlength{\tabcolsep}{16pt}
\begin{tabular}{c|c|c} 
\specialrule{1.2pt}{0pt}{0pt}
\multirow{2}{*}{} & \multicolumn{2}{c}{Average (CIR)}  \\ 
\cline{2-3}
                  & PSNR   & SSIM                       \\ 
\hline
IRNeXt~\cite{cui2024revitalizing}            & 32.568 & 0.957                      \\
IRNeXt w/ CFE     & 32.823 & 0.958                      \\
\hline
Restormer~\cite{zamir2022restormer}         & 31.621 & 0.950                      \\
Restormer w/ CFE  & 31.813 & 0.950                      \\
\specialrule{1.2pt}{0pt}{0pt}
\end{tabular}
\label{tab:cfeabla}
\end{table}

\subsubsection{The Kernel Size of LDO}
We further explored the impact of local dynamic convolution kernel sizes on performance and conducted a performance comparison on the deraining task. We tested the effects of three dynamic kernel sizes: $3\times3$, $5\times5$, and $7\times7$, and the results are shown in Table~\ref{tab:ks}. It can be observed that on the Rain100L, as the kernel size increases, the PSNR slightly decreases; whereas on the Rain100H, the PSNR significantly improves with the increase in kernel size, with k=7 showing an improvement of about 0.6 dB compared to k=3.

\begin{table}[!h]
\centering
\caption{The impact of different dynamic convolution kernel sizes in the LDO on single image deraining.}
\setlength{\tabcolsep}{8pt}
\begin{tabular}{c|cc|cc} 
\specialrule{1.2pt}{0pt}{0pt}
\multirow{2}{*}{kernel size} & \multicolumn{2}{c|}{Rain100L} & \multicolumn{2}{c}{Rain100H}  \\ 
\cline{2-5}
                             & PSNR   & SSIM                 & PSNR   & SSIM                 \\ 
\hline
7                            & 38.933 & 0.983                & 33.664 & 0.931                \\
5                            & 38.954 & 0.983                & 33.231 & 0.926                \\
3                            & 39.090 & 0.983                & 33.086 & 0.925                \\
\specialrule{1.2pt}{0pt}{0pt}
\end{tabular}
\label{tab:ks}
\end{table}

\subsection{Conclusion}
\label{sec:conclu}

This paper presents an efficient multiple-in-one IR model, MDIR, capable of handling various single and mixed image degradation types within a single model and parameter set. Addressing the limitations of Vanilla methods when faced with unknown or mixed degradation factors, MDIR introduces a local dynamic optimization module, optimizing both high- and low-frequency across different areas. Additionally, the model incorporates a conditional feature embedding module, leveraging degradation-specific features obtained via multi-label learning to guide the decoder in effectively identifying and utilizing related features. To validate the IR model's effectiveness, a new dataset containing both single and mixed degradation types is introduced. Experimental results demonstrate that MDIR achieves state-of-the-art performance across both mixed degradation and classical single-task scenarios, balancing efficiency and performance and establishing a new benchmark and method for image restoration research.
\clearpage
{
    \small
    \bibliographystyle{ieeenat_fullname}
    \bibliography{main}
}


\end{document}